\documentclass[times,twocolumn,final]{elsarticle}

\usepackage{medima}
\usepackage{framed,multirow}

\usepackage{amssymb}
\usepackage{latexsym}

\usepackage{url}
\usepackage{xcolor}
\usepackage{hyperref}
\usepackage{graphicx}
\usepackage{algorithm,algorithmic}
\usepackage{amsmath,amssymb,amsfonts}
\usepackage{textcomp}
\usepackage{booktabs}
\usepackage{multirow} 
\usepackage{amsmath,amssymb,amsfonts}
\usepackage{float}
\definecolor{newcolor}{rgb}{.8,.349,.1}
\journal{Medical Image Analysis}

\hypersetup{hidelinks} 

\begin{document}

\verso{X. Zhang \textit{et~al.}}

\begin{frontmatter}

\title{HELPNet: Hierarchical Perturbations Consistency and Entropy-guided Ensemble for Scribble Supervised Medical Image Segmentation}%

\author[1]{Xiao \snm{Zhang}\fnref{t1}}
\author[1]{Shaoxuan \snm{Wu}\fnref{t1}}
\fntext[t1]{Xiao Zhang and Shaoxuan Wu contribute equally.}
\cortext[t2]{Corresponding author.}
\author[1]{Peilin \snm{Zhang}}
\author[1]{Zhuo \snm{Jin}}
\author[2]{Xiaosong \snm{Xiong}}
\author[1]{Qirong \snm{Bu}}
\author[3]{Jingkun \snm{Chen}\corref{t2}}
\ead{jingkun.chen@eng.ox.ac.uk}
\author[1]{Jun \snm{Feng}\corref{t2}}
\ead{fengjun@nwu.edu.cn}

\address[1]{School of Information Science and Technology, Northwest University, Xi'an, China}
\address[2]{School of Biomedical Engineering \& State Key Laboratory of Advanced Medical Materials and Devices, ShanghaiTech University, Shanghai, China}
\address[3]{Institute of Biomedical Engineering, Department of Engineering Science, University of Oxford, Oxford, UK}

\received{x x xxxx}
\finalform{x x xxxx}
\accepted{x x xxxx}
\availableonline{x x xxxx}

\begin{abstract}
Creating fully annotated labels for medical image segmentation is prohibitively time-intensive and costly, emphasizing the necessity for innovative approaches that minimize reliance on detailed annotations. Scribble annotations offer a more cost-effective alternative, significantly reducing the expenses associated with full annotations. However, scribble annotations offer limited and imprecise information, failing to capture the detailed structural and boundary characteristics necessary for accurate organ delineation.
To address these challenges, we propose HELPNet, a novel scribble-based weakly supervised segmentation framework, designed to bridge the gap between annotation efficiency and segmentation performance. HELPNet integrates three modules. The Hierarchical perturbations consistency (HPC) module enhances feature learning by employing density-controlled jigsaw perturbations across global, local, and focal views, enabling robust modeling of multi-scale structural representations. Building on this, the Entropy-guided pseudo-label (EGPL) module evaluates the confidence of segmentation predictions using entropy, generating high-quality pseudo-labels. Finally, the structural prior refinement (SPR) module incorporates connectivity and bounded priors to enhance the precision and reliability and pseudo-labels. Experimental results on three public datasets ACDC, MSCMRseg, and CHAOS show that HELPNet significantly outperforms state-of-the-art methods for scribble-based weakly supervised segmentation and achieves performance comparable to fully supervised methods. The code is available at \href{https://github.com/IPMI-NWU/HELPNet}{https://github.com/IPMI-NWU/HELPNet}.
\end{abstract}


\end{frontmatter}

\section{Introduction}
\label{sec1}
\begin{figure}[!t]
\centerline{\includegraphics[width=\columnwidth]{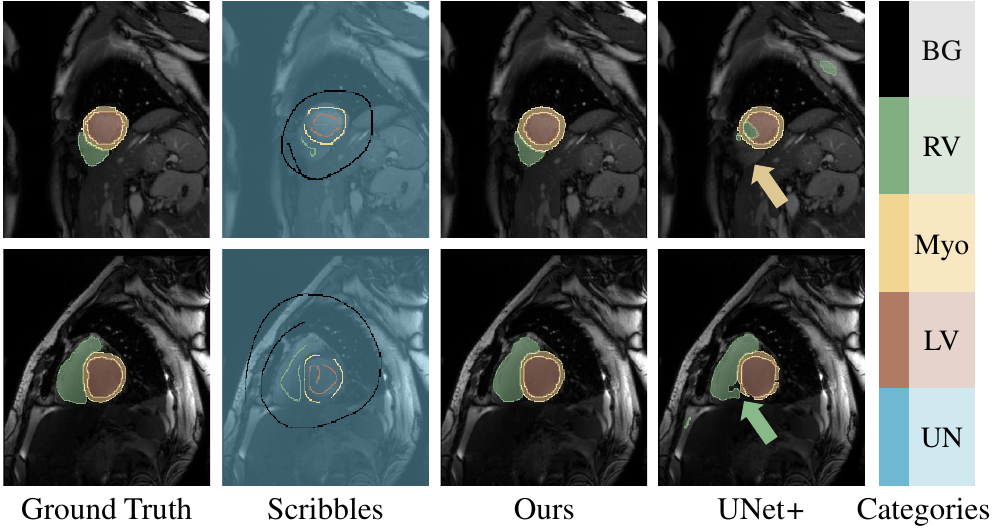}}
\caption{Two cases highlight the challenges of scribble supervision. Intricate internal structures and limited boundaries cause the baseline (UNet$+$) to inaccurately preserve or even lose the RV shape, as indicated by the arrows.
In contrast, our method significantly outperforms by effectively addressing these challenges. BG, RV, Myo, LV, and UN represent the background, right ventricle, myocardium, left ventricle, and unannotated pixels, respectively.}
\label{fig:fig0}
\end{figure}

Deep learning models have exhibited exceptional performance in medical image segmentation tasks \citep{recent_Intro,zhang2023anatomy,chen2024dynamic}. 
However, achieving competitive accuracy and robust generalization typically depends on the availability of extensive fully annotated data.
The manual annotation of medical images, which demands significant expertise from professional clinicians, is time-intensive and costly. To mitigate this limitation, weakly supervised learning methods have been employed for medical image segmentation. 
These methods leverage weak annotations, such as image-level labels \citep{imagelevel3,imagelevel4}, point annotations \citep{point2}, bounding box annotations \citep{bbox3,bbox2} and scribble annotations \citep{CycleMix,ScribFormer}, to train models with reduced annotation effort \citep{COVID19}.
Among these, scribble annotations are particularly advantageous due to their versatility and ability to assist the annotation of complex organs in medical images \citep{review1}.
As shown in Fig.\ref{fig:fig0}, scribble annotations require drawing only a few representative lines or strokes for each category, making them a practical and efficient alternative to full annotations. 

Despite their advantages, scribbles as supervision come with several challenges. Their sparse and localized nature weakens the supervision signal, particularly in low-contrast regions within the target's internal substructures. This often leads to misclassifications and constrains the overall segmentation performance of the baseline model, as highlighted by the yellow arrows in Fig.\ref{fig:fig0}.
Additionally, Unlike full annotation, scribbles provide only partial and incomplete labeling, which restricts their ability to capture target shapes and boundaries. This limitation hinders the ability of the model to preserve target morphology accurately, as illustrated by the boundary errors of the baseline model (a modified UNet, denoted as UNet+) \citep{CycleMix} shown with green arrows in Fig.\ref{fig:fig0}.

To address challenges in scribble supervision, existing methods have made numerous attempts to assist networks in acquiring shape information.
Adversarial learning frameworks, which aim to capture shape priors through generative modeling, typically rely on additional fully annotated data \citep{ACCL,MAAG}. Meanwhile, consistency regularization focuses on maintaining stability in segmentation predictions under transformations such as geometric rotations \citep{PacingPseudo}, cutout \citep{ShapePU}, or mixup operations \citep{CycleMix}. 
By leveraging invariance of shape features across various augmentations, consistency regularization facilitates learning of shape priors \citep{CycleMix, ShapePU}. 
However, these methods mainly concentrate solely on image-level consistency, while overlooking fine-grained local details, which reduces their ability to capture complex structures and limits their overall performance.

The jigsaw puzzle technique, extensively used in self-supervised learning, serves as an effective image enhancement method for enhancing image understanding \citep{Jigsaw1,Jigsaw2,Jigsaw3}. 
By dividing an image into patches and randomly shuffles them randomly, it disrupts the spatial layout while preserving semantic content, encouraging the network to learn spatial structures and relationships within the image.
To enhance the network's ability to learn multi-scale structural representations and intricate organ shapes, we propose the Hierarchical Perturbations Consistency (HPC) module. This module integrates a multi-level feature extraction strategy, systematically analyzing targets from three perspectives: global view (No Jigsaw), local view (Low-Density Jigsaw), and focal view (High-Density Jigsaw), as shown in Fig.\ref{fig:fig1}. No jigsaw perturbation facilitate the learning of global context, low-density jigsaw perturbation enhance the extraction of local structural features, and high-density jigsaw perturbation focus on capturing fine-grained, localized details \citep{Jigsaw4}. By enforcing consistency across predictions generated under varying levels of jigsaw perturbation, the module can learn a comprehensive representation of target characteristics at multiple granularities. Specifically, it aligns the segmentation results of jigsaw-perturbed and restored images with those of the original image, compelling the network to accurately model complex structures and spatial relationships. This process not only enhances network's ability to capture both global context and intricate structural details but also strengthens its robustness to perturbations.

To enhance supervision and reduce boundary errors, pseudo-labels (PL) techniques are typically applied to generate signals for unlabeled pixels \citep{DMPLS,Pseudo1}. However, existing methods rely on combining multiple predictions through mean or randomly weighted summation, which introduces noise caused by uncertainty pedictions and compromises the reliability of the generated PLs \citep{DMPLS,TDNet,ScribFormer}.
To address this limitation, we propose the entropy-guided PL ensemble (EGPL), which evaluates the confidence of the segmented prediction maps using information entropy under varying perturbation densities.  By assigning weights to prediction maps based on their entropy values and combining them accordingly, EGPL generates more reliable pseudo-labels that capture target structures.
Furthermore, structural prior PL refinement (SPR) is introduced to eliminate noise and enhance PL quality. SPR incorporates connectivity and boundary information of human organs as prior knowledge, effectively refining PL by removing noise and delineating accurate target boundaries through connectivity analysis and real-time edge detection models.

In this paper, we propose a novel scribble-guided weakly supervised medical image segmentation framework, HELPNet. The framework significantly improves the ability of the network to learn global and local information about complex organs by introducing HPC, EGPL, and SPR modules. HPC facilitates the learning of comprehensive feature representation through varying densities of jigsaw puzzles, thereby improving the network's understanding of target structures. EGPL employs information entropy to assess the confidence of segmentation results at different scales, generating high-quality PL. Additionally, SPR utilizes connectivity and edge information to further enhance the accuracy and reliability of PL.

The contributions of this paper are as follows:
1) We propose HELPNet, which leverages hierarchical perturbation and entropy-guided PL ensembles to effectively learn global and local features, achieving state-of-the-art performance on three public datasets.
2) The HPC module incorporates jigsaw perturbations into weakly supervised learning, enabling the extraction of multiscale feature representations across three granularities—global, local, and focal views—to enhance the comprehension of complex target structures.
3) The EGPL module ensemble high-quality PL from the predictions of hierarchical perturbations, while the SPR module further enhances the accuracy and reliability of PL by utilizing connectivity and edge information.

\section{Related Works}
\subsection{Scribble-based Medical Image Segmentation}
Scribble-based annotations have emerged as a promising approach for medical image segmentation, offering a cost-effective alternative to traditional pixel-wise annotations. By marking sparse curves on target regions, scribble annotations significantly reduce annotation effort while providing sufficient information for weakly supervised learning \citep{BrainTumor,CycleMix,DMSPS,ScribFormer}. 
However, the sparse and localized nature of scribble annotations introduces challenges. Their limited supervisory signal and coverage of target regions restrict the network's ability to accurately capture object shapes and boundaries, leading to suboptimal segmentation performance.
Various methods address this issue using consistency learning \citep{Transconsistent,ShapePU} and pseudo-labels (PL) \citep{Scribble2Label,DMPLS}.
For example, CycleMix \citep{CycleMix} maximizes supervision by using the mix augmentation strategy for global and local consistency to improve performance. 
\citet{DMPLS} generates PL by dynamically mixing the predictions of two decoders using random weights, which increases diversity. 
\citet{ScribFormer} combines the predictive probabilities from CNN and Transformer to generate PL, thereby supervising unlabeled pixels and reducing errors.
In this work, we introduce the concept of hierarchical jigsaw into scribble supervision. Different densities of jigsaw puzzles promote different granularity of feature learning by the network, which improves the network's ability to understand the structure of the target.

Consistency learning is crucial in scribble-guided segmentation\citep{BortsovaConsistency,CycleMix,chen2023semi}. The principle is that transformed images should have corresponding transformed predictions. Ensuring consistency across various augmentations aids the network in comprehending the target structure.
\citet{BortsovaConsistency} proposed a twin-network architecture with two identical branches, each processing differently transformed versions of the same image and trained to produce consistent predictions. 
\citet{Transconsistent} introduced the consistency of the transformation, improving the generalization of the model by aligning the segmentation results with the input transformations (\emph{e.g.}, rotation, flipping). 
\citet{CycleMix} employed a mix augmentation strategy and cycle consistency to enhance medical image segmentation performance. 
ShapePU \citep{ShapePU} applied Cutout to crop images, requiring consistent predictions post-cropping to learn shape information. 
However, such methods focus only on image-level consistency, ignoring the impact of multi-scale perturbation on boosting the model's ability to learn about complex structures.
To address this, we introduce the hierarchical jigsaw, which maintains semantic information while enabling the network to learn the multi-scale target structure.

\subsection{Pseudo Labels in Scribble-guided Segmentation}
Pseudo Labels (PL) enhance supervision and compensate for scribble-guided segmentation by using a model trained on labeled data to predict labels for unlabeled data \citep{DMPLS, S2ME, PCLMix}. Typically, PL is generated by integrating the outputs of different networks or branches.
For instance, 
\citet{DMPLS} dynamically mix predictions of two decoders, increasing the PL diversity and mitigating weaknesses. 
\citet{CompetetoWin} select the high confident pixels from the confidence maps of multiple networks as PL.  
\citet{TDNet} uses a three-branch network with different dilation rates, averaging feature-level perturbations to obtain PL.
PacingPseudo \citep{PacingPseudo} generates PL through a twin network architecture, where the predictions of one network serve as consistency targets for the other, increasing the robustness.
\citet{HidesClass} extract class labels from scribble and integrate them into PL using class activation maps. 

Current PL methods are typically average or randomly weight multiple prediction maps, which leads to introducing noise caused by uncertain predictions and reducing the quality of the PL \citep{wang2021uncertainty}.
Therefore, we propose using entropy to evaluate the confidence of hierarchical perturbation predictions and assign weights for the ensemble, resulting in a more reliable PL of higher quality.

\begin{figure*}[ht]
\centerline{\includegraphics[width=18cm]{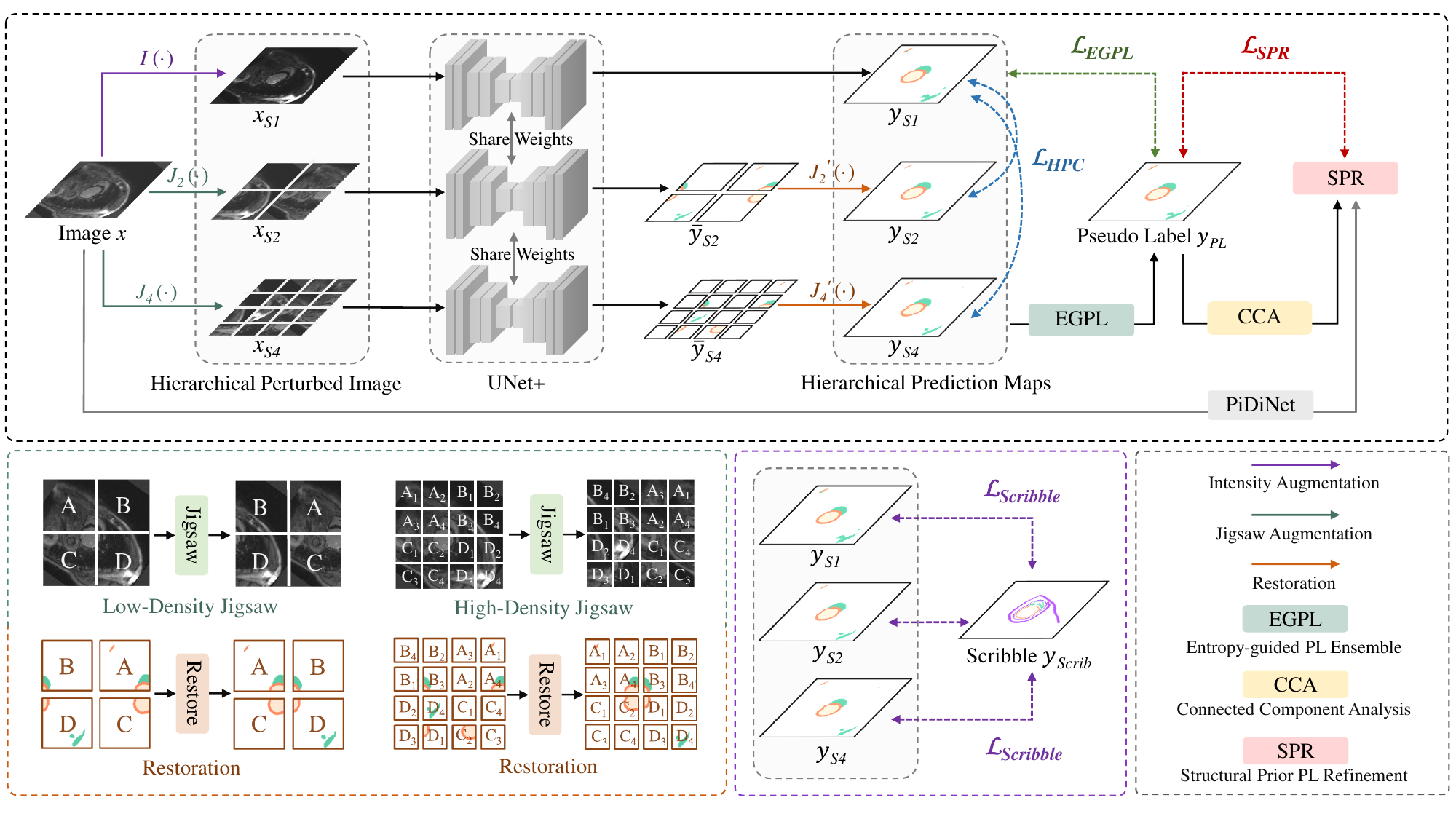}}
\caption{Overview of the proposed HELPNet that employs hierarchical jigsaw perturbations to generate predictions at various scales. Entropy-guided PL is ensembled to enhance supervision signals. Further, Structural priors are subsequently utilized to refine PL.}
\label{fig:fig1}
\end{figure*}

\section{Method}
Fig.\ref{fig:fig1} illustrates the framework of our proposed HELPNet, which consists of three key components. 
Firstly, the HPC module intensifies and restructures input images by applying jigsaw perturbations of varying densities, creating a hierarchy of perturbed images. These perturbed images are processed by three shared-weight segmenters, which generate predictions and simultaneously perform restoration. The consistency observed across the three predicted maps facilitates the network's acquisition of both local details and global structural information. Secondly, the EGPL module assesses the confidence of predictions using information entropy and allocates weights accordingly to ensemble reliable PL. 
Finally, the SPR module analyzes connected components of PL and extracts precise image boundaries, further refining these labels' quality. 

\subsection{Hierarchical Perturbation Consistency}
The limited information provided by scribble annotations regarding precise structural shapes poses a challenge for network aiming to maintain accurate segmentation capabilities. To address this, we propose the HPC module, which leverages hierarchical perturbed images and consistency learning to facilitate the network to capture feature information at three granular levels: global, local, and focal.

Specifically, given an input image $x$, to enrich the diversity of the data, we first apply intensity enhancement $I(\cdot)$ to obtain the brightness and contrast enhanced image $x_{S1}$. 
Subsequently, to direct the network's attention towards local information, we employ low-density jigsaw augmentation (\emph{i.e.,} $2\times2$ patches) $J_2(\cdot)$ and high-density jigsaw augmentation (\emph{i.e.,} $4\times4$ patches) $J_4(\cdot)$ to perturb the original image $x$, resulting in $x_{S2}$ and $x_{S4}$, respectively. Perturbed images can be enhanced as follows: 
\begin{equation}
    x_{S1} = I(x), ~~ x_{S2} = J_2(x),  ~~ x_{S4} = J_4(x).
    \label{eq_1}
\end{equation}

The hierarchical perturbations described above preserve semantic information while disrupting spatial structures. The varying densities of the jigsaw guide the network's attention from global to more local feature information. After hierarchical perturbations, the images are fed into each of the three weight-sharing segmenters UNet$+$, resulting in three prediction maps: $y_{S1}$, $\bar{y}_{S2}$, and $\bar{y}_{S4}$. Since the jigsaw process disrupts spatial positions, we need to restore $\bar{y}_{S2}$ and $\bar{y}_{S4}$, which can be formulated as: 
\begin{equation}
    {y}_{S2} = J'_{2}(\bar{y}_{S2}),~~{y}_{S4} = J'_{4}(\bar{y}_{S4}),
    \label{eq_2}
\end{equation}
where $J'_{2}$ and $J'_{4}$ are inverse functions of $J_{2}$ and $J_{4}$ respectively.
The concept of HPC mandates that the three predicted values remain consistent to aid the network in learning intricate structure information. The consistency loss can be expressed as: 
\begin{equation}
    \mathcal{L}_{HPC} = \mathcal{L}_{cos}({y}_{S2}, {y}_{S1}) + \mathcal{L}_{cos}({y}_{S4}, {y}_{S1}).
    \label{eq_3}
\end{equation}
Through alignment, the network harmonizes the shapes learned globally and locally, facilitating a deeper understanding of target shapes at different granularities. Here, $\mathcal{L}_{cos}$ represents the cosine similarity distance, measuring the similarity between two vector spaces. It can be written as: 
\begin{equation}
\mathcal{L}_{cos}(y_{S2}, y_{S1}) = 1 -  \frac{y_{S2} \cdot y_{S1}} {||y_{S2}||_{2} \cdot ||y_{S1}||_{2}}.
\label{eq_4}
\end{equation}

\begin{figure*}[ht]
\centerline{\includegraphics[width=16cm]{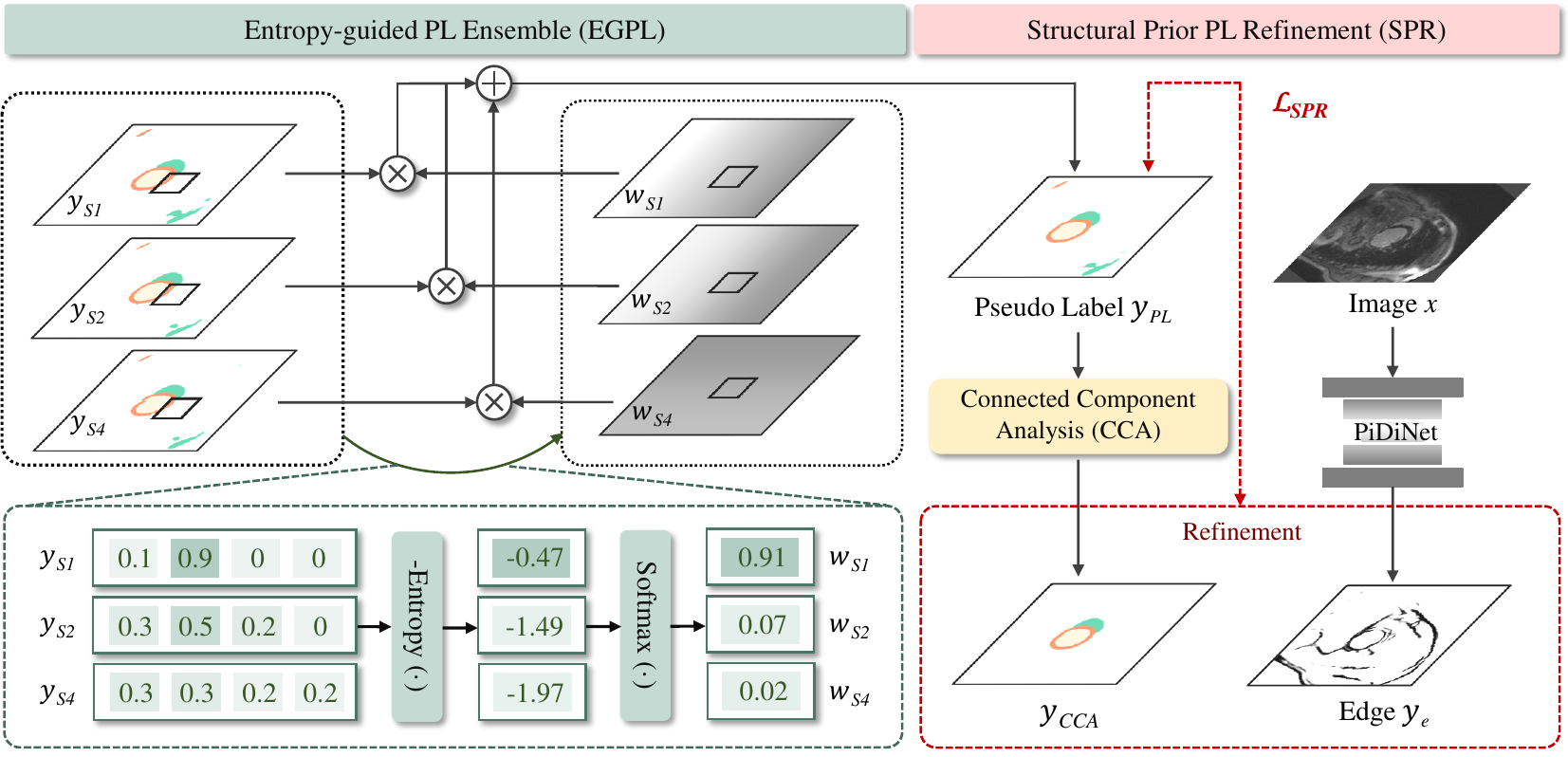}}
\caption{Entropy is employed by the entropy-guided PL to assess the confidence of predictions and ensemble PL. Structural priors further refine PL through connected component analysis and edge detection.}
\label{fig:fig2}
\end{figure*}

\subsection{Entropy-guided Pseudo Label Ensemble}
To fully exploit the potential of unlabeled region, PL techniques provide a means to enrich supervision information, thereby compensating for the limited and inadequate scribble supervision area. However, conventional methods, such as simple averaging, indiscriminately assign equal or random weights to each predicted map, rendering them vulnerable to noise, particularly in the boundary region of the organ. In this section, we present the EGPL ensemble module, which assesses the confidence of diverse prediction maps using information entropy and subsequently assigns appropriate weights to synthesize PL.

Specifically, as shown in Fig.\ref{fig:fig2}, we aggregate hierarchical prediction maps to construct the $y_{PL}$, which can be formulated as:
\begin{equation}
    y_{PL} = argmax(y_{S1} \cdot w_{S1} + y_{S2} \cdot w_{S2} + y_{S4} \cdot w_{S4}),
\label{eq_5}
\end{equation}
where $y_{S1}$, $y_{S2}$, and $y_{S4}$ represent the three prediction maps with dimensions $H \times W \times C$. $H$ and $W$ are the height and width of the image, and $C$ represents the number of segmentation classes. The weights $w_{S1}$, $w_{S2}$, and $w_{S4}$ for these prediction maps are all of dimension $H \times W \times 1$. The weight assignment is expressed as: 
\begin{equation}
    [w_{S1}^{i,j}, w_{S2}^{i,j}, w_{S4}^{i,j}] = \mathcal{S}([-E(y_{S1}^{i,j}), -E(y_{S2}^{i,j}), -E(y_{S4}^{i,j})]),
\label{eq_6}
\end{equation}
where $i$ and $j$ represent the coordinates in the length and width, respectively. $E(\cdot)$ denotes the entropy function, defined as: \begin{equation}
    E(y^{i,j}) = - \textstyle \sum_{k=1}^{C}y^{i,j,k}log_{2}y^{i,j,k}, 
\end{equation}
$E(\cdot)$ quantifies the confidence of prediction maps. Smaller entropy values indicate higher network confidence, leading to greater weight during aggregation. As shown in Eq.\ref{eq_6}, the three entropies are activated using the softmax function $\mathcal{S}(\cdot)$, ensuring their sum equals 1. We incorporate a temperature coefficient $\tau$ in the softmax function to scale the results, emphasizing weight differences, which is defined as:
\begin{equation}
    \mathcal{S}(x) = \frac{e(x_i/\tau)}{ {\textstyle \sum_{i=1}^{K}e(x_i/\tau)} }.
\label{eq_7}
\end{equation}

Once we obtain the PL $y_{PL}$, we utilize it for supervising the three prediction maps $\Omega_y = \left\{{y_{S1}, y_{S2}, y_{S4}}\right\}$ by Dice loss. The calculation equation is as follows:
\begin{align}
   \mathcal{L}_{EGPL} = \sum_{y\in \Omega_y}^{} \mathcal{L}_{Dice}(y, y_{PL}), 
\label{eq_8}
\end{align}
where $\mathcal{L}_{Dice}$ represents the Dice loss, quantifying the overlap between PL and hierarchical prediction maps, \emph{i.e.}, $\mathcal{L}_{Dice}(y_1, y_2) = 1 - 2(y_1 \cap y_2) / (|y_1|+|y_2|) $.

\subsection{Structural Prior Pseudo Label Refinement}
Given the inherent limitations of scribble annotations, particularly their lack of complete edge and shape information, accurately segmenting boundaries remains a formidable challenge for networks. Consequently, incorporating structural prior knowledge to assist the network in learning organ shapes becomes imperative. In this context, we propose the SPR module, which leverages both organ connectivity and image boundary information as priors to enhance the original PL.

As illustrated in Fig.\ref{fig:fig2}, our method commences by employing connected component analysis to eliminate non-maximal connected regions for each class within the PL, resulting in the refined PL denoted as $y_{CCA}$. This step effectively mitigates noise interference. Simultaneously, we harness an existing lightweight edge extraction algorithm called PiDiNet \citep{PiDiNet} to extract edge information $y_e$ from the original images. Notably, PiDiNet is pre-trained on BSDS500 \citep{BSDS500} and its FPS is as high as 253.
The loss function for the SPR module can be formulated as follows: 
\begin{equation}
    \mathcal{L}_{SPR} = \mathcal{L}_{cos}(y_{PL}, y_{CCA}) + \mathcal{L}_{Edge}(\hat{y}_{PL}, y_{e}),
\label{eq_9}
\end{equation}
where $\hat{y}_{PL}$ represents the foreground probability values of the PL before the argmax operation. Its dimensions are $1 \times H \times W$, and it can be expressed as $\hat{y}_{PL} = \sum_{c=1}^{K}y_{PL}^{c}$.
$\mathcal{L}_{Edge}$ is defined as:
\begin{equation}
    \mathcal{L}_{Edge} = \frac{\sum_{i=1}^{H}\sum_{j=1}^{W}
    \mathbb{I}(y^{i,j}_{e}> \alpha )||\hat{y}_{PL}^{i,j}-y^{i,j}_{e}||^2} {\sum_{i=1}^{H}\sum_{j=1}^{W}\mathbb{I}(y^{i,j}_{e}>\alpha )},
\label{eq_10}
\end{equation}
where $i,j$ represents the coordinates of $y$. The indicator function $\mathbb{I}(\cdot)$ evaluates to 1 when the probability exceeds the threshold and 0 otherwise. Specifically, when the probability exceeds $\alpha$, set here at 0.5, the point is considered part of the boundary region and requires the PL's foreground probability at that point must be similar.

\subsection{Scribble Supervision}
To leverage information from scribble annotations, we supervise the hierarchical prediction maps using partial cross-entropy loss (pce). The loss function is defined as follows:
\begin{equation}
    \mathcal{L}_{Scribble} = \sum_{y\in \Omega_y}^{} \mathcal{L}_{pce}(y, y_{Scrib})
\label{eq_11}
\end{equation}
where $y_{Scrib}$ represents the one-hot scribble labels and $\mathcal{L}_{pce}$ denotes the partial cross-entropy loss, which is defined as:
\begin{equation}
    \mathcal{L}_{pce}(p,y) = 
    \sum_{i\in \Omega_s}^{}
    \sum_{c\in K}y^{i}_c\log p^i_c,
\label{eq_12}
\end{equation}
where, $\Omega_s$ represents the coordinate set of annotated pixels in the scribble. 

The final optimization objective is given by:
\begin{equation}
    \mathcal{L} =  \mathcal{L}_{scribble} + \lambda_1 \mathcal{L}_{HPC} + \lambda_2 \mathcal{L}_{EGPL} + \lambda_3\mathcal{L}_{SPR}
\label{eq_13}
\end{equation}

\section{Experiments}
\subsection{Datasets and Evaluation Metrics}
Our method was validated using three publicly available datasets, MSCMRseg \citep{MSCMR1,MSCMR2}, ACDC \citep{ACDC}, and CHAOS \citep{CHAOS}.
The MSCMRseg dataset comprises late gadolinium enhancement cardiac MR images from 45 cardiomyopathy patients. Published annotations, including the gold standard and scribble annotations \citep{CycleMix}, are available for the right ventricle (RV), left ventricle (LV), and myocardium (Myo) in these MR images. Per \citep{CycleMix}, the 45 patients were divided into 25 cases for training, 5 cases for validation, and 15 cases for testing.
The ACDC dataset consists of 2D cine-MR images from 100 patients, acquired using 1.5T or 3T MR scanners with varying temporal resolutions. ACDC dataset includes annotated end-diastolic and end-systolic phase images for each patient \citep{ACDC}, with gold standard and scribble annotations for RV, LV, and Myo. Following \citep{CycleMix}, the 100 patients were split into 35 cases for training, 15 cases for validation, and 15 cases for testing, excluding the remaining 35 for fair comparison.
The CHAOS dataset provides T1-DUAL and T2-SPIR MRI sequences for 20 patients \citep{CHAOS}, with masks and scribbles for the liver (LIV), right kidney (RK), left kidney (LK), and spleen (SPL). 
Training and evaluation were conducted separately on the two modalities \citep{MAAG, PacingPseudo}. The Dice coefficient was employed to evaluate the performance of various methods.

\subsection{Experimental Details}
All experiments were conducted using PyTorch and executed on NVIDIA 3080Ti GPU. Initially, the intensity of each image was normalized to have a zero mean and unit variance. To enhance the training data, random flips, rotations, and noise were applied to the images. Subsequently, each image was sized to $224\times224$ pixels for input into the network. The Adam optimizer was employed to minimize the loss function Eq.\ref{eq_13}, with a decay rate of 0.95. The batch size, epochs, and learning rate were set to 4, 1000, and 0.0001, respectively. Based on empirical evidence and experimental results, the weights  $\lambda_1, \lambda_2, $ and $ \lambda_3$ in Eq.\ref{eq_13} were assigned values of $0.3, 0.1, $ and $ 0.3$, respectively. For comparative purposes with other methods, during the inference phase, the predictions obtained from the segmenter on the original images were used as the final results, without any additional optimization strategies.

\begin{table}[!t] 
    \centering
    \caption{ 
    The Dice score of our method on the MSCMRseg dataset is compared with various other methods. Bold indicates the best performance among weakly supervised methods. \\
    }
    \resizebox{\linewidth}{!}{
    \renewcommand\arraystretch{1.14}
    \setlength{\tabcolsep}{2.3pt}
    \scriptsize
    \begin{tabular}{l|lll|l}
    \toprule[0.5pt]
    
    Method    & LV  & MYO  & RV  & Avg   \\
    
    \midrule[0.3pt]

    \textcolor{blue}{$\spadesuit$} UNet$+$ \citep{CycleMix}
    & .494\tiny$\pm$.08  & .583\tiny$\pm$.06
    & .057\tiny$\pm$.02  & .378 \\

    \textcolor{blue}{$\spadesuit$} UNet$++$
    \citep{ScribFormer}
    & .497 & .472 & .506 & .492 \\

    \textcolor{blue}{$\spadesuit$} Puzzle Mix
    \citep{Puzzle}
    & .061\tiny$\pm$.02  & .634\tiny$\pm$.08 
    & .028\tiny$\pm$.01  & .241 \\

    \textcolor{blue}{$\spadesuit$} Co-mixup
    \citep{Comixup}
    & .356\tiny$\pm$.08  & .343\tiny$\pm$.07 
    & .053\tiny$\pm$.02  & .251 \\

    \textcolor{blue}{$\spadesuit$} MixUp
    \citep{mixup}
    & .610\tiny$\pm$.14  & .463\tiny$\pm$.15 
    & .378\tiny$\pm$.15  & .484 \\
    
    \textcolor{blue}{$\spadesuit$} Cutout
    \citep{Cutout}
    & .459\tiny$\pm$.08  & .641\tiny$\pm$.13
    & .697\tiny$\pm$.15  & .599 \\

    \textcolor{blue}{$\spadesuit$} CutMix
    \citep{CutMix}
    & .578\tiny$\pm$.06  & .622\tiny$\pm$.12
    & .761\tiny$\pm$.11  & .654 \\

    \textcolor{blue}{$\spadesuit$}  CycleMix
    \citep{CycleMix}
    & .870\tiny$\pm$.06  & .739\tiny$\pm$.05
    & .791\tiny$\pm$.07  & .800 \\

    \textcolor{blue}{$\spadesuit$}  DMPLS
    \citep{DMPLS}
    & .898\tiny$\pm$.08  & .791\tiny$\pm$.12
    & .798\tiny$\pm$.19  & .829 \\

    \textcolor{blue}{$\spadesuit$}  ScribFormer
    \citep{ScribFormer}
    & .896 & .813  & .807  & .839 \\

    \textcolor{blue}{$\spadesuit$}  ShapePU  
    \citep{ShapePU}
    & .919\tiny$\pm$.03  & .832\tiny$\pm$.04
    & .804\tiny$\pm$.12  & .852 \\

     \textcolor{blue}{$\spadesuit$}  \textbf{Ours}  
    & \textbf{.932}\tiny$\pm$.04  & \textbf{.861}\tiny$\pm$.04
    & \textbf{.896}\tiny$\pm$.04  & \textbf{.896} \\

    \midrule[0.3pt]

    \textcolor{red}{$\blacksquare$} UNet$_{F}$
    \citep{UNet}
    & .850 & .738 & .721 & .770 \\

    \textcolor{red}{$\blacksquare$} UNet$+_{F}$
    \citep{CycleMix}
    & .857\tiny$\pm$.06  & .720\tiny$\pm$.08
    & .689\tiny$\pm$.12  & .755 \\

    \textcolor{red}{$\blacksquare$} UNet$++_{F}$
    \citep{UNet++}
    & .866  & .731  & .745  & .774 \\

    \textcolor{red}{$\blacksquare$} Puzzle Mix$_F$
    \citep{Puzzle}
    & .867\tiny$\pm$.04  & .742\tiny$\pm$.04
    & .759\tiny$\pm$.04  & .789 \\

    \textcolor{red}{$\blacksquare$} CycleMix$_F$
    \citep{CycleMix}
    & .864\tiny$\pm$.03  & .785\tiny$\pm$.04
    & .781\tiny$\pm$.07  & .810 \\

    \textcolor{red}{$\blacksquare$} nnUNet
    \citep{nnUNet}
    & .944  & .882
    & .880  & .902 \\
   
    \bottomrule[0.5pt]
    \end{tabular}
    }
    \label{tab:tab1}
\end{table}

\begin{figure*}[ht]
\centerline{\includegraphics[width=\textwidth]{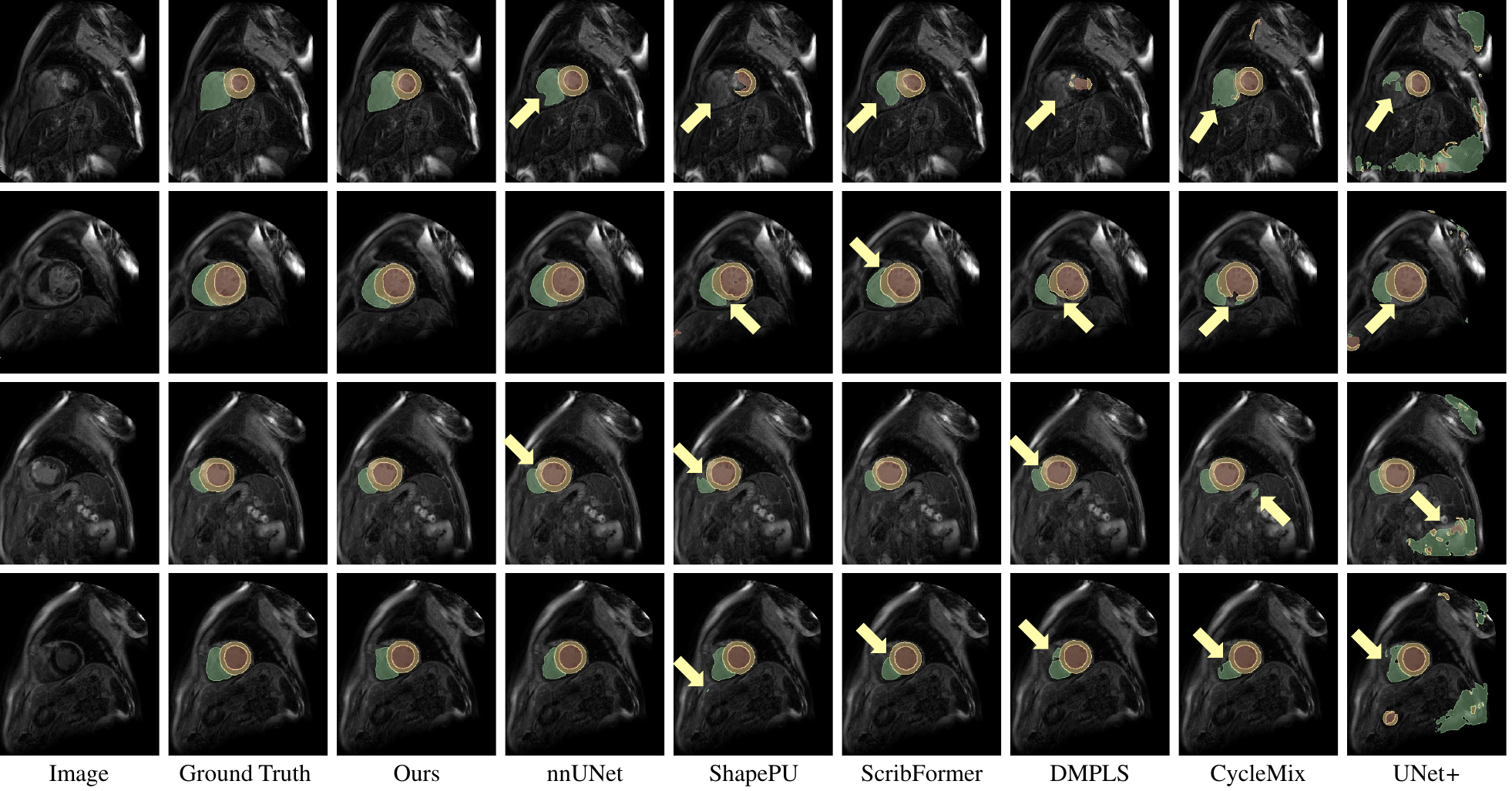}}
\caption{Visualization on four typical cases from
MSCMRseg dataset for illustration and comparison. The yellow arrows highlight mis-segmentations}
\label{fig:fig3}
\end{figure*}

\subsection{Results on MSCMRseg Dataset}
Table.\ref{tab:tab1} presents a comparative analysis of our method's performance against other methods on the MSCMRseg dataset. Two types of supervision were evaluated: scribble supervision (blue spade \textcolor{blue}{$\spadesuit$}) and full supervision (red square \textcolor{red}{$\blacksquare$}). 
Methods such as PostDAE \citep{PostDAE} necessitate additional full annotation, and given the scarcity of MSCMRseg, these methods were not included in the comparison. Within the scribble supervision category, the following methods were compared:
1) UNet and its variants: Includes UNet$+$ \citep{CycleMix} and UNet$++$ \citep{UNet++}.
2) Different augmentation strategies on UNet$+$, including Puzzle Mix \citep{Puzzle}, Co-mixup \citep{Comixup}, Mixup \citep{mixup}, Cutout \citep{Cutout}, CutMix \citep{CutMix}, and CycleMix \citep{CycleMix}.
3) Other Scribble Segmentation Frameworks that utilize PL or consistency mechanisms, including DMPLS \citep{DMPLS} with dynamic mixed PL, ScribFormer \citep{ScribFormer}, and ShapePU \citep{ShapePU} with cutout consistency learning.
In the full supervision category, comparisons were made with several UNet-related networks, such as UNet$_{F}$ \citep{UNet}, UNet$+_{F}$ \citep{CycleMix}, UNet$++_{F}$ \citep{UNet++}, and nnUNet \citep{nnUNet}. Additionally, fully supervised forms of semi-supervised frameworks were included. Except for some unreported experimental results, other methods are presented as mean $\pm$ standard deviation.

As illustrated in Table.\ref{tab:tab1}, our method significantly outperformed other weakly supervised methods, achieving an average Dice score of 89.6\%, which is 4.4\% higher than the previous best method, ShapePU.
Compared to the CycleMix method, which also employs consistency, HELPNet outperforms it by 9.6\% in average Dice score (89.6\% vs. 80.0\%). Additionally, when compared to the DMPLS method, which also utilizes PL, HELPNet demonstrates a 6.7\% improvement in average Dice score (89.6\% vs. 82.9\%).
In the segmentation of the RV, our method demonstrated an 8.9\% improvement compared to ScribFormer (89.6\% vs. 80.4\%). 
Moreover, our method achieves performance comparable to the current state-of-the-art fully supervised method, nnUNet (89.6\% vs. 90.2\% in average Dice), and outperforms it on RV (89.6\% vs. 88.0\%).

Fig.\ref{fig:fig3} visualizes the qualitative experiments of our method compared to other state-of-the-art methods on the MSCMRseg dataset. Our segmentation results are the most accurate, closely aligning with the ground truth and exhibiting fewer outlier segmentation regions. In the segmentation of the RV (green area), our method performed exceptionally well. As shown in the first row, other methods produced unsatisfactory results for the RV, displaying shape inconsistency (ScribFormer, nnUNet), disconnection (CycleMix), or even organ loss (ShapePU, DMPLS, UNet$+$), marked with yellow arrows. Additionally, in the results of Myo (yellow area) and LV (red area), our results highly overlapped with the gold standard. Other methods exhibited varying degrees of under-segmentation and over-segmentation. Excellent segmentation performance underscores the effectiveness of our approach. The HPC module significantly aids the network in learning the multi-scale structures and intricate shape details, while the EGPL module and SPR module provides richer signals for training and accurate target boundaries.

\begin{figure*}[ht]
\centerline{\includegraphics[width=\textwidth]{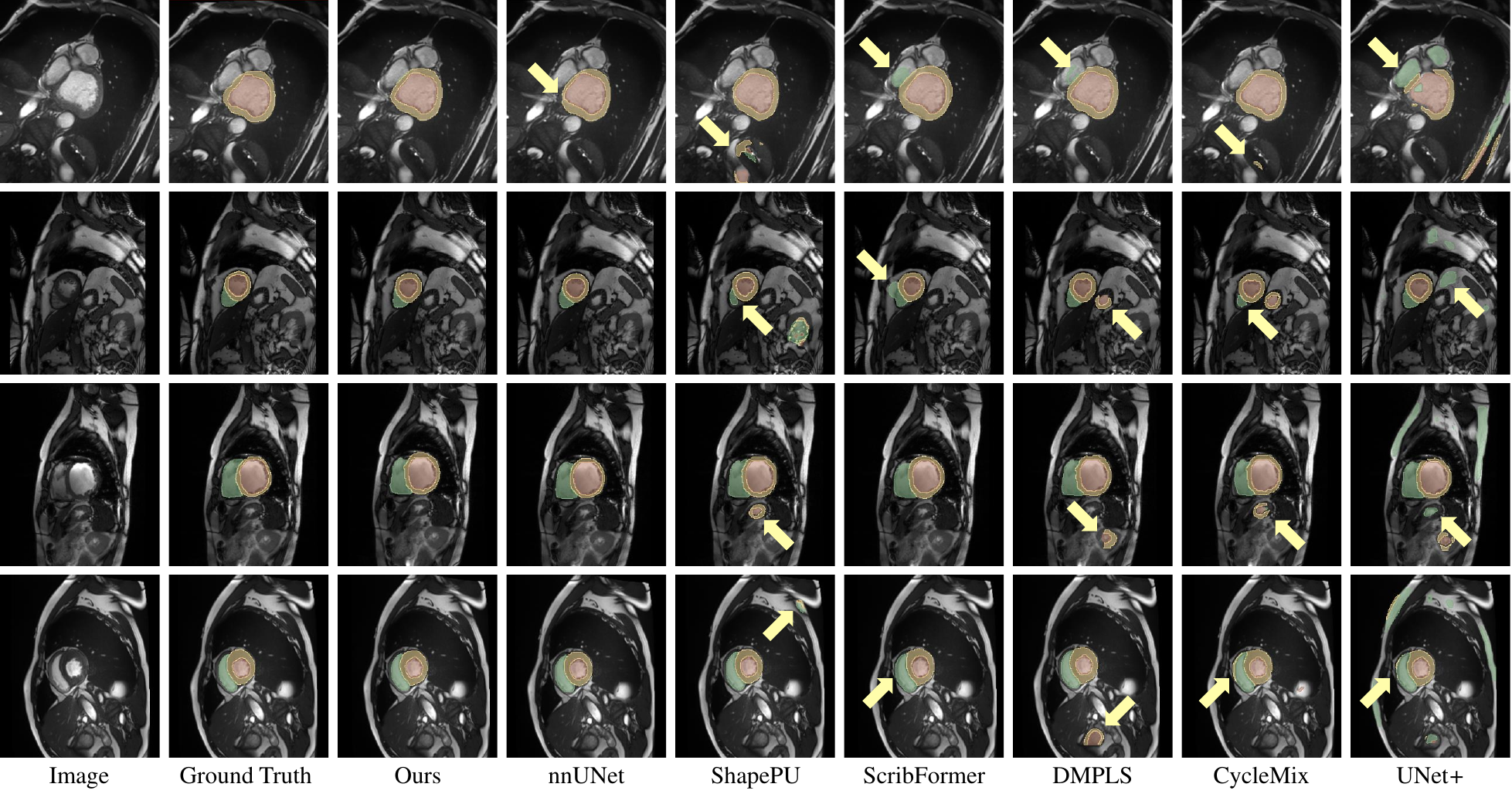}}
\caption{
Qualitative comparison between HELPNet and other state-of-the-art methods on ACDC dataset. Network mis-segmentation is indicated by yellow arrows.
}
\label{fig:fig4}
\end{figure*}

\begin{table}[!]
    \centering
    \caption{ 
    The Dice score of our method on the ACDC dataset is compared with various other methods. Bold indicates the best performance among weakly supervised methods.\\
    }
    \resizebox{\linewidth}{!}{
    \renewcommand\arraystretch{1.14}
    \setlength{\tabcolsep}{2.3pt}
    \scriptsize
    \begin{tabular}{l|lll|l}
    \toprule[0.5pt]
    
    Method    & LV  & MYO  & RV  & Avg   \\
    
    \midrule[0.3pt]
    
    \textcolor{blue}{$\spadesuit$} UNet
    \citep{UNet}
    & .842\tiny$\pm$.07  & .764\tiny$\pm$.06
    & .693\tiny$\pm$.11  & .766 \\

    \textcolor{blue}{$\spadesuit$} UNet$+$
    \citep{CycleMix}
    & .785\tiny$\pm$.20  & .725\tiny$\pm$.15
    & .746\tiny$\pm$.20  & .752 \\

    \textcolor{blue}{$\spadesuit$} UNet$++$
    \citep{UNet++}
    & .846 & .652 & .787 & .761 \\

    \textcolor{blue}{$\spadesuit$} UNet$_{wpce}$
    \citep{wpce}
    & .784\tiny$\pm$.09  & .675\tiny$\pm$.06
    & .563\tiny$\pm$.13  & .674 \\

    \textcolor{blue}{$\spadesuit$} UNet$_{crf}$
    \citep{CRF}
    & .766\tiny$\pm$.09  & .661\tiny$\pm$.06
    & .590\tiny$\pm$.14  & .672 \\

    \textcolor{blue}{$\spadesuit$} Puzzle Mix
    \citep{Puzzle}
    & .663\tiny$\pm$.33  & .650\tiny$\pm$.23 
    & .559\tiny$\pm$.34  & .624 \\

    \textcolor{blue}{$\spadesuit$} Co-mixup
    \citep{Comixup}
    & .622\tiny$\pm$.30  & .621\tiny$\pm$.21 
    & .702\tiny$\pm$.21  & .648 \\

    \textcolor{blue}{$\spadesuit$} MixUp
    \citep{mixup}
    & .803\tiny$\pm$.18  & .753\tiny$\pm$.12 
    & .767\tiny$\pm$.23  & .774 \\
    
    \textcolor{blue}{$\spadesuit$} Cutout
    \citep{Cutout}
    & .832\tiny$\pm$.17  & .754\tiny$\pm$.14
    & .812\tiny$\pm$.13  & .800 \\

    \textcolor{blue}{$\spadesuit$} CutMix
    \citep{CutMix}
    & .641\tiny$\pm$.36  & .734\tiny$\pm$.14
    & .740\tiny$\pm$.22  & .705 \\
    
    \textcolor{blue}{$\spadesuit$} PacingPseudo
    \citep{PacingPseudo}
    & .777\tiny$\pm$.29  & .825\tiny$\pm$.17
    & .884\tiny$\pm$.19  & .829 \\

    \textcolor{blue}{$\spadesuit$}  CycleMix    
    \citep{CycleMix}
    & .883\tiny$\pm$.10  & .798\tiny$\pm$.08
    & .863\tiny$\pm$.07  & .848 \\

    \textcolor{blue}{$\spadesuit$}  DMPLS
    \citep{DMPLS}
    & .861\tiny$\pm$.10  & .842\tiny$\pm$.05
    & \textbf{.913}\tiny$\pm$.08  & .872 \\

    \textcolor{blue}{$\spadesuit$}  ScribFormer
    \citep{ScribFormer}
    & .922 & .871  & .871  & .888 \\

    \textcolor{blue}{$\spadesuit$}  ShapePU  
    \citep{ShapePU}
    & .860\tiny$\pm$.12  & .791\tiny$\pm$.09
    & .852\tiny$\pm$.10  & .834 \\

     \textcolor{blue}{$\spadesuit$}  \textbf{Ours}  
    & \textbf{.933}\tiny$\pm$.06  & \textbf{.900}\tiny$\pm$.02
    & .884\tiny$\pm$.09  & \textbf{.906} \\

    \midrule[0.3pt]

    \textcolor{green}{$\blacklozenge$} PostDAE
    \citep{PostDAE}
    & .806\tiny$\pm$.07  & .667\tiny$\pm$.07
    & .556\tiny$\pm$.12  & .675 \\

    \textcolor{green}{$\blacklozenge$} UNet$_D$
    \citep{MAAG}
    & .753\tiny$\pm$.09  & .597\tiny$\pm$.08
    & .404\tiny$\pm$.15  & .584 \\

     \textcolor{green}{$\blacklozenge$} ACCL
     \citep{ACCL}
    & .878\tiny$\pm$.06  & .797\tiny$\pm$.05
    & .735\tiny$\pm$.10  & .803 \\

     \textcolor{green}{$\blacklozenge$} MAAG
     \citep{MAAG}
    & .879\tiny$\pm$.05  & .817\tiny$\pm$.05
    & .752\tiny$\pm$.12  & .816 \\

    \midrule[0.3pt]    

    \textcolor{red}{$\blacksquare$} UNet$_{F}$
    \citep{UNet}
    & .892 & .789 & .830 & .837 \\

    \textcolor{red}{$\blacksquare$} UNet$+_{F}$
    \citep{CycleMix}
    & .883\tiny$\pm$.13  & .831\tiny$\pm$.09
    & .870\tiny$\pm$.10  & .862 \\

    \textcolor{red}{$\blacksquare$} 
    UNet$++_{F}$
    \citep{UNet++}
    & .875  & .771  & .798  & .815 \\

    \textcolor{red}{$\blacksquare$} SwinUnet
    \citep{SwinUnet}
    & .900  & .818  & .812  & .843 \\

    \textcolor{red}{$\blacksquare$} UNETR
    \citep{UNETR}
    & .926  & .845  & .844  & .872 \\

    \textcolor{red}{$\blacksquare$} Puzzle Mix$_F$
    \citep{Puzzle}
    & .912\tiny$\pm$.08  & .842\tiny$\pm$.08
    & .887\tiny$\pm$.07  & .880 \\

    \textcolor{red}{$\blacksquare$} CycleMix$_F$
    \citep{CycleMix}
    & .919\tiny$\pm$.07  & .858\tiny$\pm$.06
    & .882\tiny$\pm$.09  & .886 \\

    \textcolor{red}{$\blacksquare$} nnUNet
    \citep{nnUNet}
    & .939\tiny$\pm$.05  & .918\tiny$\pm$.02
    & .900\tiny$\pm$.03  & .917 \\
   
    \bottomrule[0.5pt]
    \end{tabular}
    }
    \label{tab:tab2}
\end{table}

\subsection{Results on ACDC Dataset}
Table.\ref{tab:tab2} presents a comparative analysis of our method's performance with other state-of-the-art on the ACDC dataset. Three types of supervision were evaluated: scribble supervision (blue spade \textcolor{blue}{$\spadesuit$}), unpaired masks and scribbles (green diamond \textcolor{green}{$\blacklozenge$}), and full supervision (red square \textcolor{red}{$\blacksquare$}). Scribble supervision methods include:
1) UNet and its variants: UNet \citep{UNet}, UNet$+$ \citep{CycleMix}, UNet$++$ \citep{UNet++}, UNet with weighted pce (UNet$_{wpce}$) \citep{wpce}, and UNet with conditional random field (UNet$_{crf}$) \citep{CRF}.
2) Different augmentation strategies on UNet$+$.
3) Other scribble-based frameworks with PL or consistency, including PacingPseudo \citep{PacingPseudo}, DMPLS \citep{DMPLS}, and ScribFormer \citep{ScribFormer} using PL, and ShapePU \citep{ShapePU} using consistency.
Additionally, methods employing extra unpaired masks were compared, such as PostDAE \citep{PostDAE}, UNet$_D$ \citep{MAAG}, ACCL \citep{ACCL}, and MMAG \citep{MAAG}. Full supervision method include the UNet series, including UNet$_F$\citep{UNet}, UNet$+_{F}$\citep{CycleMix}, UNet$++_{F}$\citep{UNet++}, SwinUnet\citep{SwinUnet}, UNETR\citep{UNETR}, and nnUNet\citep{nnUNet}. Fully supervised forms of weakly supervised frameworks were also included.
The results in Table.\ref{tab:tab2} demonstrate that our method outperformed other weakly supervised methods, achieving an average Dice score of 90.6\% on the ACDC dataset, which is 1.8\% higher than the previous best method, ScribFormer. Our method achieved the best results in RV and Myo segmentation, ranks second in RV segmentation, behind DMPLS. Although DMPLS excels with large organs using dynamic pseudo labels, it struggles with detailed areas like Myo.
In the second part of the Table.\ref{tab:tab2}, comparing with methods using unpaired data, our method also outperformed others. 
Finally, in comparison with fully supervised methods, our method outperformed most methods and exhibited performance on par with the state-of-the-art nnUNet.

\begin{table*}[!ht] \scriptsize
    \centering
    \caption{ 
    The Dice score of our method on the CHAOS T1\&T2 dataset is compared with various other methods. Bold indicates the best performance among weakly supervised methods.\\
    }
    \resizebox{0.98\linewidth}{!}{
    \renewcommand\arraystretch{1.14}
    \setlength{\tabcolsep}{6.4pt}
    \scriptsize
    \begin{tabular}{l|lllll|lllll}
    \toprule[0.5pt]
    
    \multirow{2}{*}{Method}    & \multicolumn{5}{c|}{T1}
    & \multicolumn{5}{c}{T2}\\
    \cmidrule {2 - 11}
    & LIV  & RK  & LK  & SPL & Avg & LIV  & RK  & LK  & SPL & Avg \\
    
    \midrule[0.3pt]
    
    \textcolor{blue}{$\spadesuit$} UNet
    \citep{UNet}
    & .435\tiny$\pm$.07  & .213\tiny$\pm$.04
    & .091\tiny$\pm$.03  & .259\tiny$\pm$.07
    & .250
    & .484\tiny$\pm$.08  & .239\tiny$\pm$.05
    & .097\tiny$\pm$.02  & .277\tiny$\pm$.07
    & .274
    \\

    \textcolor{blue}{$\spadesuit$} UNet$_{wpce}$
    \citep{wpce}
    & .425\tiny$\pm$.09  & .292\tiny$\pm$.02
    & .166\tiny$\pm$.02  & .257\tiny$\pm$.05
    & .285
    & .556\tiny$\pm$.09  & .315\tiny$\pm$.04
    & .284\tiny$\pm$.03  & .322\tiny$\pm$.10
    & .370\\

    \textcolor{blue}{$\spadesuit$} UNet$_{crf}$
    \citep{CRF}
    & .373\tiny$\pm$.09  & .200\tiny$\pm$.06
    & .163\tiny$\pm$.04  & .279\tiny$\pm$.13
    & .254 
    & .480\tiny$\pm$.09  & .264\tiny$\pm$.15
    & .199\tiny$\pm$.03  & .329\tiny$\pm$.12
    & .318 \\

    \textcolor{blue}{$\spadesuit$}  DMPLS
    \citep{DMPLS}
    & .519\tiny$\pm$.36  & .498\tiny$\pm$.31
    & .469\tiny$\pm$.34  & .225\tiny$\pm$.28
    & .428
    & .484\tiny$\pm$.38  & .645\tiny$\pm$.29
    & .585\tiny$\pm$.33  & .217\tiny$\pm$.30
    & .484 \\
    
    \textcolor{blue}{$\spadesuit$} PacingPseudo
    \citep{PacingPseudo}
    & \textbf{.805}\tiny$\pm$.27  & .665\tiny$\pm$.33
    & .636\tiny$\pm$.36  & .613\tiny$\pm$.36
    & .680 
    & .791\tiny$\pm$.30  & \textbf{.787}\tiny$\pm$.30
    & .771\tiny$\pm$.32  & .598\tiny$\pm$.42
    & \textbf{.737} \\

    \textcolor{blue}{$\spadesuit$} \textbf{Ours}
    & .781\tiny$\pm$.11  & .614\tiny$\pm$.12
    & \textbf{.725}\tiny$\pm$.04  & \textbf{.633}\tiny$\pm$.11
    & \textbf{.688} 
    & \textbf{.793}\tiny$\pm$.06  & .736\tiny$\pm$.11
    & \textbf{.808}\tiny$\pm$.06  
    & \textbf{.602}\tiny$\pm$.17
    & .735 \\

    \midrule[0.3pt]

    \textcolor{green}{$\blacklozenge$} PostDAE
    \citep{PostDAE}
    & .328\tiny$\pm$.07  & .579\tiny$\pm$.07
    & .571\tiny$\pm$.06  & .584\tiny$\pm$.11
    & .516 
    & .434\tiny$\pm$.07  & .579\tiny$\pm$.07
    & .575\tiny$\pm$.06  & .584\tiny$\pm$.11
    & .543 \\

    \textcolor{green}{$\blacklozenge$} UNet$_D$
    \citep{MAAG}
    & .602\tiny$\pm$.05  & .464\tiny$\pm$.10
    & .469\tiny$\pm$.06  & .413\tiny$\pm$.12 
    & .487
    & .636\tiny$\pm$.04  & .530\tiny$\pm$.10
    & .450\tiny$\pm$.08  & .341\tiny$\pm$.10 
    & .489 \\

    \textcolor{green}{$\blacklozenge$} ACCL
    \citep{ACCL}
    & .650\tiny$\pm$.12  & .573\tiny$\pm$.06
    & .494\tiny$\pm$.09  & .512\tiny$\pm$.14 
    & .557
    & .632\tiny$\pm$.10  & .428\tiny$\pm$.10
    & .465\tiny$\pm$.09  & .565\tiny$\pm$.12 
    & .523 \\

    \textcolor{green}{$\blacklozenge$} MAAG
    \citep{MAAG}
    & .640\tiny$\pm$.07  & \textbf{.685}\tiny$\pm$.06
    & .596\tiny$\pm$.09  & .397\tiny$\pm$.08 
    & .580 
    & .563\tiny$\pm$.06  & .686\tiny$\pm$.07
    & .614\tiny$\pm$.09  & .442\tiny$\pm$.08 
    & .576 \\

    \midrule[0.3pt]    
    
    \textcolor{red}{$\blacksquare$} UNet$_{F}$
    \citep{CycleMix}
    & .779\tiny$\pm$.30  & .644\tiny$\pm$.37
    & .647\tiny$\pm$.38  & .531\tiny$\pm$.40 
    & .650 
    & .778\tiny$\pm$.33  & .752\tiny$\pm$.35
    & .653\tiny$\pm$.41  & .628\tiny$\pm$.42 
    & .703 \\

    \textcolor{red}{$\blacksquare$} UNet$_{FD}$
    \citep{PacingPseudo}
    & .786\tiny$\pm$.30  & .659\tiny$\pm$.37
    & .674\tiny$\pm$.39  & .558\tiny$\pm$.40 
    & .670
    & .779\tiny$\pm$.34  & .745\tiny$\pm$.36
    & .676\tiny$\pm$.40  & .649\tiny$\pm$.42 
    & .712 \\

    
    \bottomrule[0.5pt]
    \end{tabular}
    }
    \label{tab:tab3}
\end{table*}

Fig.\ref{fig:fig4} presents the qualitative visualization of our HELPNet on the ACDC dataset. Our segmentation results closely align with the ground truth, demonstrating the accurate capture of the target shape, comparable to the fully supervised nnUNet. In the example in the first row, nnUNet's segmentation of the Myo is less accurate than our method. 
The ACDC dataset contains many regions with similar contrast and structures to the atria, leading to misclassification by other methods such as DMPLS and ShapePU.
However, our proposed HPC module focuses on both global, local and focal views, learning from multiple levels of information about the complex structure of the target.
Due to the lack of shape information in scribble annotations, some weakly supervised methods struggle to maintain segmentation accuracy, as seen with CycleMix in the examples in the second and fourth rows. 
In contrast, our proposed SPR module optimizes the network to maintain precise shape using organ connectivity and boundaries. 
Consequently, our method demonstrates superior performance without over-segmentation.

\subsection{Results on CHAOS Dataset}
Table.\ref{tab:tab3} presents a comparative analysis of the performance of our method against others on CHAOS T1 and T2, respectively. 
We compare with three types of methods: scribble supervision (blue spade \textcolor{blue}{$\spadesuit$}), unpaired masks and scribbles (green diamond \textcolor{green}{$\blacklozenge$}), and full supervision (red square \textcolor{red}{$\blacksquare$}).
UNet$_{FD}$ indicates a fully supervised UNet using cross-entropy loss and dice loss.
On CHAOS T1, our method achieved the highest Dice score, significantly exceeding DMPLS, which also employs PL (68.8\% vs. 42.8\%). The result demonstrates the capability of our method to generate high-quality PL through entropy weighting.
Compared to PacingPseudo, HELPNet shows superior performance in terms of the average Dice score. On the CHAOS T2 dataset, HELPNet also performed exceptionally well, comprehensively outperforming all methods that use additional labeled data (PostDAE, UNet$_D$, etc.) on both T1 and T2. Furthermore, our method matches the performance of fully supervised methods.

\begin{table}[t]
    \centering
    \caption{ 
    Ablation studies were conducted on the MSCMRseg dataset under different settings to investigate the contribution of various components. Symbol $^\ast$ indicates statistically significant improvement given by a Wilcoxon test with $ p < 0.05 $. Bold denotes the best performance.
    \\}
    \resizebox{\linewidth}{!}{
    \renewcommand\arraystretch{1.14}
    \setlength{\tabcolsep}{2pt}
    \scriptsize
    \begin{tabular}{cccc|ccc|c}
    \toprule[0.5pt]

     $\mathcal{L}_{Scribble}$ &  $\mathcal{L}_{HPC}$ 
    &  $\mathcal{L}_{EGPL}$     &  $\mathcal{L}_{SPR}$
    &   LV &   Myo &   RV &   Avg \\

    \midrule[0.3pt]
     \checkmark & & & 
    &.790\tiny$\pm$.09  &.766\tiny$\pm$.05
    &.771\tiny$\pm$.06  &.775 \\
    
     \checkmark & \checkmark & & 
    &~.877$^\ast$\tiny$\pm$.07  &~.811$^\ast$\tiny$\pm$.04
    &~.879$^\ast$\tiny$\pm$.04  &~.856$^\ast$ \\
    
     \checkmark & & \checkmark & 
    &~.880$^\ast$\tiny$\pm$.06  &~.809$^\ast$\tiny$\pm$.05
    &.774\tiny$\pm$.12  &~.821$^\ast$ \\

     \checkmark & & \checkmark & \checkmark
    &~.896$^\ast$\tiny$\pm$.09  &~.822$^\ast$\tiny$\pm$.06
    &~.845$^\ast$\tiny$\pm$.11  &~.855$^\ast$ \\
    
     \checkmark & \checkmark & \checkmark & 
    &~.921$^\ast$\tiny$\pm$.05  &~.841$^\ast$\tiny$\pm$.04
    &.880\tiny$\pm$.05  &~.881$^\ast$ \\
    
     \checkmark & \checkmark & \checkmark & \checkmark 
    & ~\textbf{.932$^\ast$}\tiny$\pm$.04
    & ~\textbf{.861$^\ast$}\tiny$\pm$.04
    & ~\textbf{.896$^\ast$}\tiny$\pm$.04
    & ~\textbf{.896$^\ast$} \\

    \bottomrule[0.5pt]
    \end{tabular}
    }
    \label{tab:tab4}
\end{table}

\subsection{Ablation Study}
\subsubsection{Effectiveness of Each Module of HELPNet}
Table.\ref{tab:tab4} investigates the contributions of various modules in HELPNet on the MSCMRseg dataset. When no strategies, the network achieves a Dice score of 77.5\%. Incorporating the HPC module increases the Dice score to 85.6\%, reflecting an improvement of 8.1\%, which enhancement indicates that HPC effectively aids the network in learning the multi-scale structures from various levels. 
Additionally, the inclusion of the EGPL module alone raises the Dice score to 82.1\%, demonstrating its effectiveness in introducing richer supervision during training from reliable PL. 
Further improvement is observed when SPR is added to EGPL. 
Ultimately, employing both the HPC and EGPL strategies together raises the Dice score to 88.1\%. The addition of SPR results in the highest Dice score of 89.6\%, with improvements in Dice scores and variances across all organs, demonstrating that SPR effectively leverages prior knowledge to help the network eliminate noise and maintain precise boundaries.

\begin{figure}[!t]
\centerline{\includegraphics[width=0.9\columnwidth]{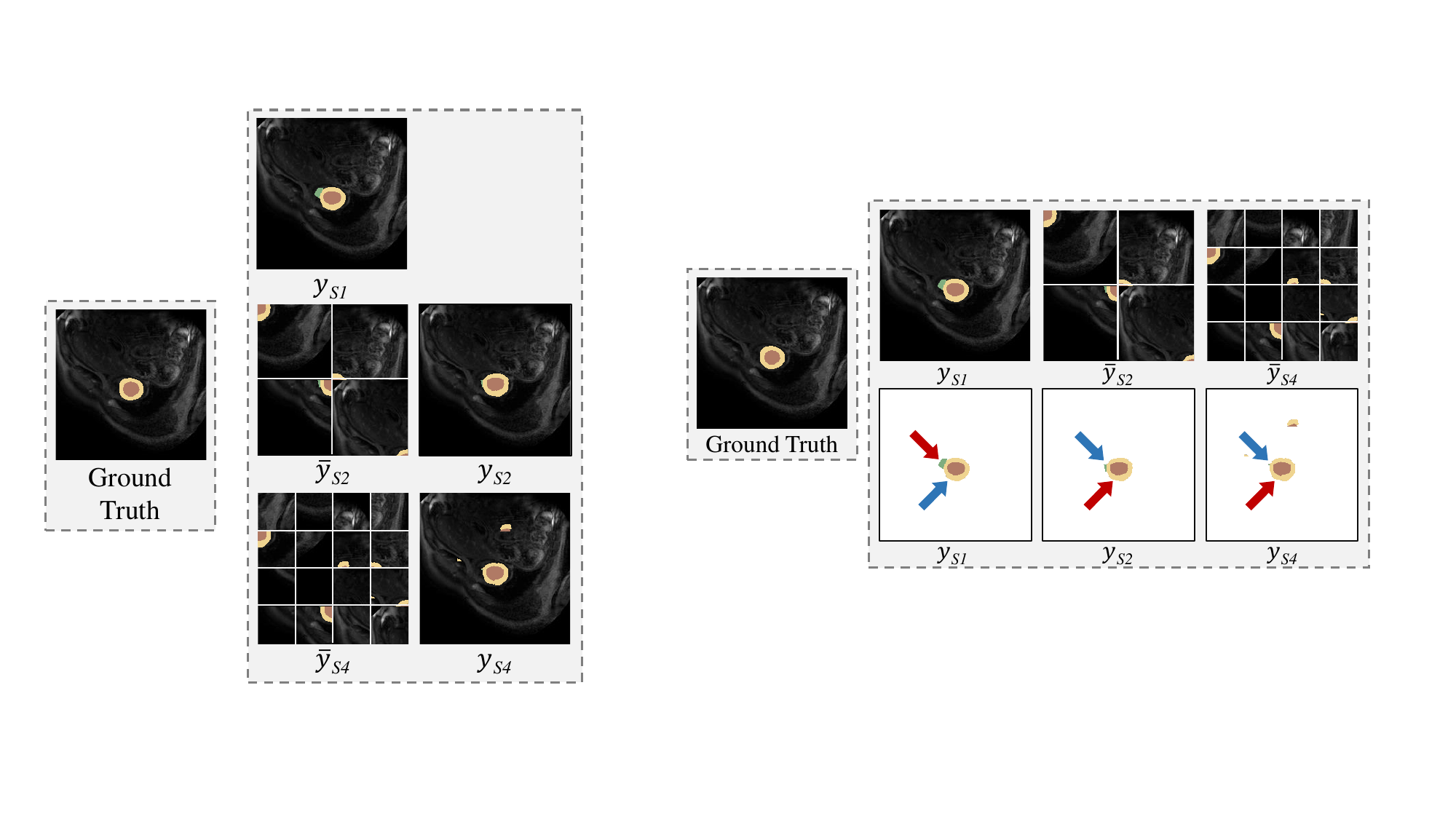}}
\caption{
Qualitative analysis of the HPC module on the MSCMRseg dataset. The red and blue arrows respectively indicate areas of segmentation errors and correct segmentation.
}
\label{fig:fig5}
\end{figure}

Fig.\ref{fig:fig5} presents the segmentation results on the MSCMRseg dataset after applying varying intensities of jigsaw perturbations. In the result of the original image $y_{S1}$, the network achieves superior overall segmentation completeness by emphasizing global information, as indicated by the blue arrow. Nevertheless, the absence of local information introduces errors, as the red arrow indicates. 
Under $y_{S2}$ for local view segmentation, the network learns the complex structure better and improves effectively on local misclassification.
Further, in the focal view's $y_{S4}$, applying a high-density puzzle perturbation allows the network to effectively eliminate segmentation errors by incorporating detailed local shape information, as shown by the blue arrows.
However, there is a deficiency in global completeness, as marked by the red arrow. 
The combination of different views enables the network to learn the multi-scale structures and complex shape details of the target.
Guided by consistency, multiple views produce complementarities that help the network learn a comprehensive representation of the target from multiple granularities.

\begin{table}[!t]
    \centering
    \caption{ 
    Ablation studies on the three weights in the loss function were performed on the MSCMRseg dataset. Bold denotes the best performance.
    \\}
    \resizebox{0.8\linewidth}{!}{
    \renewcommand\arraystretch{1.1}
    \setlength{\tabcolsep}{8pt}
    \scriptsize
    \begin{tabular}{c|ccc|c}
    \toprule[0.5pt]
    
    $\lambda$  & LV  & Myo  & RV & Avg \\
    \midrule[0.3pt]
    $\lambda_1$  & \multicolumn{4}{l}{
    $\lambda_2=0.1,\lambda_3=0.3$
    } \\
    
    \midrule[0.3pt]
    
    1 & .922\tiny$\pm$.04  & .846\tiny$\pm$.04
    & .880\tiny$\pm$.05  & .883 \\
    
    0.5 & .933\tiny$\pm$.03  & .856\tiny$\pm$.03
    & .880\tiny$\pm$.06  & .890 \\
    
    0.3 & .932\tiny$\pm$.04  & \textbf{.861}\tiny$\pm$.04
    & \textbf{.896}\tiny$\pm$.04  & \textbf{.896} \\

    0.1 & \textbf{.934}\tiny$\pm$.03  & .856\tiny$\pm$.04
    & .891\tiny$\pm$.04  & .893 \\

     \midrule[0.3pt]
    $\lambda_2$ & \multicolumn{4}{l}{
    $\lambda_1=0.3,\lambda_3=0.3$}\\ 
    \midrule[0.3pt]

    1 & .929\tiny$\pm$.04  & .856\tiny$\pm$.04
    & .882\tiny$\pm$.05  & .889 \\
    
    0.5 & .929\tiny$\pm$.04  & .857\tiny$\pm$.03
    & .886\tiny$\pm$.06  & .891 \\
    
    0.3 & \textbf{.933}\tiny$\pm$.03  & .856\tiny$\pm$.04
    & .886\tiny$\pm$.04  & .892 \\
    
    0.1 & .932\tiny$\pm$.04  & \textbf{.861}\tiny$\pm$.04
    & \textbf{.896}\tiny$\pm$.04  & \textbf{.896} \\

     \midrule[0.3pt]
    $\lambda_3$ & \multicolumn{4}{l}{
    $\lambda_1=0.3,\lambda_2=0.1$}\\ 
    \midrule[0.3pt]

    1 & .932\tiny$\pm$.03  & .855\tiny$\pm$.04
    & .819\tiny$\pm$.03  & .868 \\
    
    0.5 & .932\tiny$\pm$.03  & .858\tiny$\pm$.03
    & .889\tiny$\pm$.04  & .893 \\
    
    0.3 & .932\tiny$\pm$.04  & \textbf{.861}\tiny$\pm$.04
    & \textbf{.896}\tiny$\pm$.04  & \textbf{.896} \\
    
    0.1 & \textbf{.934}\tiny$\pm$.03  & .859\tiny$\pm$.03
    & .886\tiny$\pm$.05  & .893 \\
    
    \bottomrule[0.5pt]
    \end{tabular}
    }
    \label{tab:tab5}
\end{table}

\subsubsection{Ablation of Hyperparameters}
To examine the impact of the $\lambda$ values in Eq.\ref{eq_13} and the temperature coefficient $\tau$ in Eq.\ref{eq_7} on segment performance, a series of ablation experiments were conducted. 
Eq.\ref{eq_13} demonstrates that the weight coefficients $\lambda_1$, $\lambda_2$, and $\lambda_3$ regulate the contributions of hierarchical perturbations consistency, entropy-guided PL, and PL refinement, respectively. To explore the optimal settings for these coefficients, we conducted experiments on the MSCMRseg dataset and repeated the performance of our HELPNet under various parameter configurations in Table.\ref{tab:tab5}. Each $\lambda$ was varied from 1 to 0.1, with values of (1, 0.5, 0.3, 0.1) used for the study. The findings reveal that HELPNet achieves optimal performance when $\lambda_1$ is set to 0.3, $\lambda_2$ to 0.1, and $\lambda_3$ to 0.3. The network's performance remains notably consistent across different parameter settings, highlighting its exceptional stability.

\begin{table}[!t]
    \centering
    \caption{ 
    Ablation studies on the temperature coefficient $\tau$ were conducted on the MSCMRseg dataset. Bold denotes the best performance. \\}
    \resizebox{0.8\linewidth}{!}{
    \renewcommand\arraystretch{1.1}
    \setlength{\tabcolsep}{8pt}
    \scriptsize
    \begin{tabular}{c|ccc|c}
    \toprule[0.5pt]

    $\tau$ & LV & Myo & RV & Avg \\
    
    \midrule[0.3pt]
    
    2.0 & \textbf{.935}\tiny$\pm$.03  & .859\tiny$\pm$.04
    & .874\tiny$\pm$.06  & .889 \\ 

    1.0 & .931\tiny$\pm$.04  & .856\tiny$\pm$.04
    & .886\tiny$\pm$.05  & .891 \\

    0.8 & .930\tiny$\pm$.04  & .854\tiny$\pm$.04
    & .888\tiny$\pm$.05  & .891\\
    
    0.6 & \textbf{.935}\tiny$\pm$.03  & .859\tiny$\pm$.04
    & .889\tiny$\pm$.05  & .894 \\

    0.4 & .932\tiny$\pm$.04  & \textbf{.861}\tiny$\pm$.04
    & \textbf{.896}\tiny$\pm$.04  & \textbf{.896} \\
    
    0.2 & \textbf{.935}\tiny$\pm$.03  & .857\tiny$\pm$.04
    & .891\tiny$\pm$.05  & .894 \\
   
    \bottomrule[0.5pt]
    \end{tabular}
    }
    \label{tab:tab6}
\end{table}

In the ablation study of the hyperparameter $\tau$, multiple values (0.2, 0.4, 0.6, 0.8, 1.0, 2.0) were tested. $\tau$ adjusts the sharpness of the output probability distribution. When $\tau$ exceeds 1, the softmax output becomes smoother, reducing network confidence. Conversely, when $\tau$ is below 1, the output variance increases. Table.\ref{tab:tab6} illustrates the impact of different $\tau$ values on segmentation results. 
The lowest average Dice score occurs at $\tau = 2$, while performance peaks at $\tau = 0.4$. The network consistently demonstrates stability across various values of $\tau$. Consequently, $\tau = 0.4$ is used in other experiments.

\section{Conclusion}
In this paper, we introduce HELPNet, a novel scribble-guided weakly supervised segmentation framework. HELPNet incorporates the HPC and EGPL modules to address the challenges of insufficient shape information and supervision signals inherent in scribble supervision. 
The HPC module facilitates the network to learn feature information at three granular levels (\emph{i.e.,} global, local, and focal) through puzzles of varying densities, thereby enhancing the understanding of target structures. 
The EGPL module assesses the confidence of different perturbed segmentation results using entropy and generates high-quality PL as weights. To further refine these PL, we propose the SPR module, which leverages connectivity and boundary priors to improve their accuracy and reliability.
These interconnected strategies enable HELPNet to achieve superior segmentation results on three public datasets: MSCMRseg, ACDC, and CHAOS. Our results suggest that HELPNet provides a viable alternative to fully supervised segmentation approaches, effectively reducing annotation costs while maintaining high-quality performance. This highlights the potential of scribble annotations as a practical solution for medical image segmentation, offering a promising direction for improving the efficiency of clinical workflows.

\section*{Acknowledgments}
This work was supported in part by the National Natural Science Foundation of China (No. 62403380, 62073260), and the Natural Science Foundation of Shaanxi Province of China (No. 2024JC-YBMS-495).


\bibliographystyle{modelnames.bst}\biboptions{authoryear}

\end{document}